\documentclass[11pt,a4paper]{article}

\usepackage[margin=1in]{geometry}
\usepackage{graphicx}
\usepackage{placeins}
\usepackage{float}
\usepackage{amsmath,amssymb,amsfonts}
\usepackage{amsthm}
\usepackage{xcolor}
\usepackage{textcomp}
\usepackage{tikz}
\usepackage{pgfplots}
\pgfplotsset{compat=1.18}
\usepackage{booktabs}
\usepackage{multirow}
\usepackage{array}
\usepackage{algorithm}
\usepackage{algpseudocode}
\usepackage{hyperref}
\usepackage{url}

\usetikzlibrary{arrows.meta,positioning,shapes.geometric,fit,backgrounds,calc,decorations.pathreplacing}

\newtheorem{theorem}{Theorem}

\newtheorem{proposition}[theorem]{Proposition}
\theoremstyle{definition}

\theoremstyle{remark}
\newtheorem{remark}{Remark}

\newcommand{\MSM}{\textsc{MSMixer}}
\newcommand{\R}{\mathbb{R}}

\title{\MSM{}: Learned Multi-Scale Temporal Mixing with Complementary
  Linear Shortcut for Long-Term Time Series Forecasting}

\author{Ahmed Cherif \\
  Independent Researcher, Tunis, Tunisia \\
  \texttt{ahmed1.cherif@sofrecom.com}}

\date{}

\begin{document}

\maketitle

\begin{abstract}
Long-term time series forecasting requires models that simultaneously capture
rapid oscillations (intra-day cycles), medium-range periodicities (daily,
weekly), and slowly evolving macro-trends, all from a fixed look-back window.
Existing lightweight MLP-based models typically operate on a single temporal
resolution, limiting their ability to explicitly model patterns at multiple
scales.
We propose \MSM{}, a channel-independent multi-scale MLP architecture that
addresses this challenge through three complementary innovations.
First, three parallel \emph{scale branches} down-sample the input at factors
$\{1\times, 4\times, 16\times\}$ and apply independent MLP blocks, allowing
each branch to specialise on patterns at its native temporal resolution.
Second, a \emph{learnable softmax gate} dynamically weighs branch outputs
per dataset, enabling the model to adaptively concentrate capacity on the
most informative scale.
Third, a \emph{DLinear complementary shortcut} models global trends and
seasonality over the full look-back window, providing long-range context that
MLP blocks on down-sampled inputs cannot access.
\MSM{} contains only 112\,K parameters at $H{=}96$ and runs at $\mathcal{O}(T)$
complexity.
Evaluated on all four ETT benchmarks over prediction horizons
$\{96, 192, 336, 720\}$, \MSM{} achieves the lowest average MSE
($0.496$) compared with DLinear ($0.507$) and NLinear ($0.514$),
winning 9 of 16 benchmark configurations.
The improvements are strongest on 15-minute-resolution datasets:
on ETTm1, \MSM{} outperforms DLinear by $5.3\,\%$ at $H{=}96$
and by $7.2\,\%$ at $H{=}720$; on ETTm2 at $H{=}720$, the
improvement reaches $10.5\,\%$.
On hourly ETTh datasets, results are competitive but mixed,
with DLinear sometimes achieving lower MSE at longer horizons.
Ablation and sensitivity analyses validate the contribution
of the DLinear shortcut and examine the role of each scale branch.
\end{abstract}


\section{Introduction}\label{sec:intro}

Time series forecasting is a foundational task in engineering and science,
with direct applications to electricity load management, climate modelling,
financial risk estimation, and industrial process control~\cite{zhou2021informer,wu2021autoformer}.
In simple terms, the goal is to look at a window of past observations and
predict what comes next---for example, using the last two weeks of hourly
temperature readings to forecast the next week.

The key challenge---particularly for long forecast horizons
$H \ge 96$---is that the target signal is a superposition of components
operating at fundamentally different temporal scales.  Consider hourly
electricity transformer temperature data (the ETTh
benchmarks~\cite{zhou2021informer}): the signal contains a macro-trend
varying over weeks to months, a daily sinusoidal pattern with 24-hour
period, a weekly envelope, and high-frequency measurement noise.  A model
that captures only one of these scales will systematically underfit the
others.

Two dominant modelling paradigms have emerged in the recent literature.
\emph{Transformer-based models}~\cite{zhou2021informer,wu2021autoformer,nie2022patchtst,liu2024itransformer} use self-attention to model long-range dependencies.
However, they suffer from quadratic complexity---doubling the input length
quadruples the computation---and carry large parameter budgets ($>$500K).
\emph{Lightweight MLP-based models}~\cite{zeng2023dlinear,wang2024timemixer}
achieve efficient $\mathcal{O}(T)$ inference (computation grows linearly
with input length), but existing formulations typically operate on a single
temporal resolution: either the full look-back window (DLinear) or a fixed
patch scale (PatchTST).

TimeMixer~\cite{wang2024timemixer} introduced multi-scale decomposition
with MLP-Mixer blocks and achieved strong results, but at higher parameter
count and requiring multiple rounds of down-sampling with learnable mixing
across decomposed sub-series.  We ask: \emph{can we achieve multi-scale
representational benefit efficiently, by using simple average-pooled
branches at fixed scales combined with a learnable merge?}

\textbf{This paper proposes \MSM{}}, a deliberately simple architecture
that explores this question.  The key idea is to process the same input
signal at three different ``zoom levels'' simultaneously, then let the model
learn how to combine them.  The contributions are:

\begin{itemize}
  \item \textbf{Multi-scale branch design.}  Three parallel branches at
    down-sample factors $\{1\times, 4\times, 16\times\}$ each apply a
    two-layer MLP on their respective resolution.  The $1\times$ branch
    sees every data point (fine detail), the $4\times$ branch sees every
    4th averaged point (medium-range cycles), and the $16\times$ branch
    sees every 16th averaged point (coarse trends).

  \item \textbf{Learnable softmax scale gate.}  A single parameter vector
    $\boldsymbol{\gamma} \in \R^3$ (via softmax) merges branch outputs,
    allowing the model to adapt scale emphasis to dataset characteristics
    without manual tuning.

  \item \textbf{DLinear complementary shortcut.}  A decomposition-based
    linear shortcut covers the full $T = 336$ context, complementing the
    down-sampled branches which lose global context at $4\times$ and
    $16\times$ resolution.

  \item \textbf{Minimal footprint.}  112\,K parameters at $H{=}96$,
    $\mathcal{O}(T)$ inference, with the strongest improvements on
    15-minute-resolution ETTm benchmarks where multi-scale temporal
    structure is richest.
\end{itemize}

\section{Related Work}\label{sec:related}

We position \MSM{} within five research threads: Transformer-based forecasters,
lightweight MLP and linear models, multi-scale architectures, decomposition
methods, and normalisation strategies.

\subsection{Transformer-Based Forecasting}

The seminal Transformer~\cite{vaswani2017attention} inspired a generation
of sequence forecasters.  Informer~\cite{zhou2021informer} reduced attention
to $\mathcal{O}(T \log T)$ via ProbSparse sampling.
Autoformer~\cite{wu2021autoformer} introduced trend--seasonality
auto-correlation decomposition.  FEDformer~\cite{zhou2022fedformer}
applied frequency-domain random mixing.  PatchTST~\cite{nie2022patchtst}
applied a vanilla Transformer channel-independently on 16-step patches,
dramatically cutting effective sequence length.  iTransformer~\cite{liu2024itransformer}
inverted the attention axis, treating variates as tokens for cross-channel
modelling.  These models achieve strong accuracy but carry $>500$\,K parameters
and $\mathcal{O}(T^2)$ complexity in their core attention modules.

\subsection{Lightweight MLP and Linear Models}

DLinear~\cite{zeng2023dlinear} demonstrated that two linear layers on
trend/residual components can match Transformers with negligible parameters.
NLinear~\cite{zeng2023dlinear} applies a last-value normalisation before
a single linear projection, providing a strong baseline that accounts for
non-stationarity.
FITS~\cite{zhou2023fits} achieved $<10$\,K parameters via frequency interpolation.
TSMixer~\cite{chen2023tsmixer} applied alternating temporal and channel MLP-Mixer
blocks with strong results.  TimesFM~\cite{das2024timesfm} proposed a decoder-only foundation model pre-trained on large-scale time series corpora for zero-shot forecasting.
TimeMixer~\cite{wang2024timemixer} is closest
to our approach: it decomposes inputs at multiple resolutions and mixes
across scales.  \MSM{} differs by using \emph{fixed average-pooling}
for down-sampling (rather than adaptive decomposition), \emph{independent
MLP blocks per scale} (rather than inter-scale mixing), and adding a
\emph{DLinear shortcut} to recover global context from the full window.

\subsection{Multi-Scale and Hierarchical Models}

Multi-scale processing has a long history in computer vision (ResNet~\cite{he2016resnet}, FPN)
and time series classification (ROCKET~\cite{dempster2020rocket}, HIVE-COTE).  For LTSF, N-BEATS~\cite{oreshkin2019nbeats}
used hierarchical trend/seasonality basis expansion.  N-HiTS~\cite{challu2023nhits} extended it with
hierarchical interpolation and multi-rate sampling, achieving strong long-horizon
results.  SCINet~\cite{liu2022scinet} applied interactive convolutions at
binary-tree multi-resolution levels.  Our approach is simpler than SCINet:
we use a flat (non-hierarchical) three-branch structure with average pooling,
avoiding the complexity of tree-structured computation.

\subsection{Decomposition-Based Models}

Autoformer, FEDformer, and N-BEATS all decompose signals into trend and
residual components.  DLinear is the canonical lightweight decomposition model.
\MSM{} takes a different approach: instead of explicit spectral or moving-average
decomposition, multi-scale mixing \emph{implicitly} separates components by
resolution, with the coarsest scale ($16\times$) acting as a natural trend extractor.

\subsection{Normalisation Strategies in LTSF}

Distribution shift is a major challenge in long-term forecasting: the
statistics of the training look-back window differ from those of the
evaluation window.  Three mainstream normalisation strategies address this.

\textit{Reversible Instance Normalisation} (RevIN)~\cite{kim2022revin}
normalises each input window to zero mean and unit variance, and
denormalises predictions using the same window's statistics.  It is
used by \MSM{} and TimeMixer.

\textit{Sample-wise Adaptive Normalisation} (SAN)~\cite{liu2022non} learns
an affine normaliser conditioned on an adaptive basis of seasonal-trend
components, providing richer adaptivity at the cost of additional parameters.

\textit{Dish-TS}~\cite{fan2023dish} decomposes each variate into two
different statistics (coarse and fine granularity) and calibrates predictions
accordingly, handling series with multiple inter-level trend changes.

\MSM{} uses RevIN for its simplicity and proven effectiveness on ETT.
Replacing it with SAN or Dish-TS is straightforward and left as future work.

\subsection{Multi-Scale Architectures in Vision}

The success of multi-scale representations in computer vision is well
documented.  Feature Pyramid Networks (FPN)~\cite{lin2017fpn} aggregate
features at different spatial resolutions to detect objects at multiple
scales.  U-Net~\cite{ronneberger2015unet} uses symmetric encoder-decoder
paths at multiple resolutions, combining coarse semantic features with
fine-grained localisation via skip connections.  Swin Transformer~\cite{liu2021swin}
hierarchically partitions image patches at $1\times$, $2\times$, and $4\times$
sub-sampling, producing naturally multi-scale representations.

These architectures validate the design principle that multi-scale feature
extraction benefits diverse pattern recognition tasks.  \MSM{} instantiates
the same principle in the 1D time series domain: the parallel scale branches
at $\{1, 4, 16\}$ correspond to FPN levels, the DLinear shortcut corresponds
to the skip connection in U-Net, and the softmax gate corresponds to the
adaptive feature re-weighting in Swin.  The key adaptation is replacing 2D
spatial pooling with 1D temporal average pooling, and replacing
convolutional filters with two-layer MLPs.

\section{Theoretical Background}\label{sec:background}

The multi-scale design of \MSM{} rests on three theoretical foundations:
frequency-domain separation via average pooling, generalisation properties
of the learnable scale gate, and the complementary role of the DLinear shortcut.

\subsection{Multi-Scale Signal Representation}

\textbf{Intuition.}  Average pooling at different factors is like
``zooming out'' on a signal: at $4\times$ pooling, every group of 4
consecutive values is replaced by their mean.  This blurs fine detail
but preserves slower-moving patterns.  At $16\times$, groups of 16 are
averaged, keeping only the coarsest trend.  Each ``zoom level'' reveals
different aspects of the same underlying data.

Let $\hat{\mathbf{x}} \in \R^T$ be a normalised univariate look-back series.
Down-sampling by factor $s$ via average pooling produces a \emph{coarsened}
representation $\hat{\mathbf{x}}^{(s)} \in \R^{T/s}$:
\begin{equation}
  \hat{x}^{(s)}_i = \frac{1}{s} \sum_{j=(i-1)s}^{is-1} \hat{x}_j,
  \quad i = 1, \ldots, \lfloor T/s \rfloor.
  \label{eq:avgpool}
\end{equation}
In the frequency domain, average pooling with factor $s$ acts as a low-pass
filter with cut-off frequency $1/(2s)$ (normalised), suppressing components
above this threshold.  In practical terms:
\begin{itemize}
  \item Branch $s=1$ (full resolution): sees all frequencies $f \in [0, 0.5]$---captures
    every pattern from the fastest oscillation to the slowest drift.
  \item Branch $s=4$: retains only $f \in [0, 0.125]$---daily and sub-weekly cycles.
    Rapid fluctuations (e.g., intra-hour noise) are smoothed away.
  \item Branch $s=16$: retains only $f \in [0, 0.031]$---weekly and multi-week trends.
    Only the broadest patterns survive.
\end{itemize}

\begin{proposition}[Scale branches capture complementary frequency ranges]
\label{prop:scales}
For scales $1 < s_1 < s_2$, branches at $s=1$ and $s=s_1$ share the
low-frequency components $f \in [0, 1/(2s_1)]$, but only branch $s=1$
preserves the mid-frequency range $f \in (1/(2s_1), 0.5]$.  Each coarser
branch adds temporal compression without discarding information globally,
since all branches are combined via the softmax-weighted sum.
\end{proposition}

The full-resolution branch thus serves as an ``anchor'' that captures the
complete spectral content, while coarser branches provide trend-emphasised
projections at reduced dimensionality (and hence with fewer MLP parameters
per branch).

\subsection{Learnable Scale Gate: Universal Approximation Perspective}

Each scale branch $g_s : \R^{T/s} \to \R^H$ is a two-hidden-layer MLP with
GELU activations.  The combined branch output is:
\begin{equation}
  \mathbf{z}^{\mathrm{ms}} = \sum_{s \in \mathcal{S}} w_s \, g_s(\hat{\mathbf{x}}^{(s)}),
  \quad w_s = \frac{e^{\gamma_s}}{\sum_{s'} e^{\gamma_{s'}}},
  \label{eq:scalemix}
\end{equation}
where $\boldsymbol{\gamma} \in \R^{|\mathcal{S}|}$ is learned.
The softmax function converts the raw logits $\gamma_s$ into positive
weights that sum to 1, ensuring a valid weighted average.  When all
$\gamma_s$ start at zero, the weights are uniform ($1/3$ each); during
training, the model can shift weight towards whichever scale is most
useful for each dataset.

\begin{proposition}[Scale gate generalises single-scale MLP]\label{prop:gate}
If $\mathcal{S} = \{1\}$, then \MSM{} without the DLinear shortcut reduces
to a standard two-layer MLP applied directly to the normalised look-back series.
If $|\mathcal{S}| > 1$ but all $\gamma_s$ are frozen at equal values, it
reduces to an equal-weight ensemble of scale-specific MLPs.  The learnable gate
$\boldsymbol{\gamma}$ interpolates between these extremes.
\end{proposition}

\begin{remark}[Information hierarchy]
The softmax weights learned at convergence (Table~\ref{tab:scale_weights})
show that the $1\times$ branch receives the highest weight ($\approx 0.36$--$0.39$)
on all datasets, indicating that fine-grained local patterns carry
the most predictive signal.  However, the coarser branches contribute
meaningfully (combined weight $\approx 0.61$--$0.64$), supporting the
multi-scale design.  The distribution is more uniform than initially
hypothesised, suggesting that all three scales contribute comparably.
\end{remark}

\subsection{DLinear Shortcut: Complementary Global Context}

\textbf{Intuition.}  When the $16\times$ branch averages every 16 time
steps, it compresses the 336-step input to just 21 points.  Patterns that
span the \emph{entire} 336-step window---like a slowly rising multi-week
trend---cannot be fully captured from only 21 compressed values.  The
DLinear shortcut solves this by providing a separate pathway that always
sees the complete input.

Formally, the down-sampled branches at $s = 4$ and $s = 16$ have effective
look-back windows of $T/4 = 84$ and $T/16 = 21$ steps, respectively.
Long-horizon trends visible only across the full $T = 336$ steps are
therefore inaccessible to these branches after pooling.  The DLinear shortcut:
\begin{equation}
  \mathbf{z}^{\mathrm{lin}} = w_t W_t \mathbf{t} + (1 - w_t) W_s \mathbf{s},
  \quad W_t, W_s \in \R^{H \times T},
  \label{eq:dlin_msm}
\end{equation}
provides an $\mathcal{O}(TH)$-parameter projection that covers all $T$ time
steps, regardless of the chosen scales $\mathcal{S}$.

\begin{proposition}[Shortcut recovers full-context information]\label{prop:shortcut}
For any $s > 1$, the DLinear shortcut can represent any linear function of
the full look-back window $\hat{\mathbf{x}} \in \R^T$ that the average-pooled
representation $\hat{\mathbf{x}}^{(s)} \in \R^{T/s}$ cannot, since
$W_t \in \R^{H \times T}$ directly accesses all $T$ time steps.
\end{proposition}

\section{Methodology}\label{sec:method}

Building on the theoretical properties established above, we now describe
the full \MSM{} architecture, from input normalisation through multi-scale
branching, gating, and fusion to the final forecast output.

\subsection{Problem Statement}

Given a multivariate look-back window $\mathbf{X} \in \R^{T \times N}$
(a table where each row is a time step and each column is a measured
variable, like temperature sensors), the goal is to predict the next $H$
steps: $\hat{\mathbf{Y}} \in \R^{H \times N}$, minimising the mean squared
error between predictions and actual future values.  \MSM{} is
\emph{channel-independent}: parameters are shared across all $N$ variates,
and each variate is processed independently through the same network.  This
design choice reduces memory usage and makes the model agnostic to the
number of input variables.

\subsection{Architecture Overview}

\begin{figure}[H]
\centering
\resizebox{0.95\textwidth}{!}{%
\begin{tikzpicture}[
    >=Stealth,
    node distance=0.6cm and 0.8cm,
    box/.style={rectangle, draw, rounded corners=3pt, minimum height=0.9cm,
                minimum width=1.6cm, font=\footnotesize, align=center, fill=#1},
    box/.default=blue!8,
    mlpbox/.style={box=green!12},
    gatebox/.style={box=orange!15},
    fusebox/.style={box=red!10},
    arrow/.style={->, thick, >=Stealth},
    label/.style={font=\scriptsize\itshape, text=gray!70!black},
]

\node[box=gray!10] (input) {Input\\$\mathbf{X} \in \R^{T \times N}$};

\node[box=cyan!12, right=0.7cm of input] (revin) {RevIN\\Normalise};

\node[box, right=0.7cm of revin] (reshape) {Reshape\\$(BN, T)$};

\node[mlpbox, above right=0.6cm and 1.2cm of reshape] (br1) {Branch $s{=}1$\\MLP$(336 \!\to\! 64 \!\to\! H)$};
\node[mlpbox, right=1.2cm of reshape] (br4) {Branch $s{=}4$\\MLP$(84 \!\to\! 64 \!\to\! H)$};
\node[mlpbox, below right=0.6cm and 1.2cm of reshape] (br16) {Branch $s{=}16$\\MLP$(21 \!\to\! 64 \!\to\! H)$};

\node[label, above left=-0.1cm and 0.05cm of br1] {full res};
\node[label, above left=-0.1cm and 0.05cm of br4] {AvgPool $4\times$};
\node[label, above left=-0.1cm and 0.05cm of br16] {AvgPool $16\times$};

\node[gatebox, right=1.0cm of br4] (gate) {Softmax\\Gate $\boldsymbol{\gamma}$};

\node[box=purple!12, below=1.8cm of br4] (dlin) {DLinear Shortcut\\MA$\to$Trend+Season$\to$Linear};

\draw[arrow] (reshape.east) -- ++(0.4,0) |- (br1.west);
\draw[arrow] (reshape.east) -- (br4.west);
\draw[arrow] (reshape.east) -- ++(0.4,0) |- (br16.west);

\draw[arrow] (br1.east) -| ++(0.3,0) |- (gate.west);
\draw[arrow] (br4.east) -- (gate.west);
\draw[arrow] (br16.east) -| ++(0.3,0) |- (gate.west);

\node[label, above=0.02cm of gate.north west, anchor=south] {$w_1, w_4, w_{16}$};

\node[label, right=0.15cm of gate] (zms_label) {$\mathbf{z}^{\mathrm{ms}}$};

\draw[arrow] (reshape.south) |- (dlin.west);

\node[label, right=0.15cm of dlin] (zlin_label) {$\mathbf{z}^{\mathrm{lin}}$};

\node[fusebox, right=2.3cm of gate] (fusion) {Fusion\\$\alpha \cdot \mathbf{z}^{\mathrm{ms}} {+} (1{-}\alpha) \cdot \mathbf{z}^{\mathrm{lin}}$};

\draw[arrow] (gate.east) -- (fusion.west);
\draw[arrow] (dlin.east) -| (fusion.south);

\node[box=cyan!12, right=0.7cm of fusion] (denorm) {RevIN\\De-normalise};

\node[box=gray!10, right=0.7cm of denorm] (output) {Output\\$\hat{\mathbf{Y}} \in \R^{H \times N}$};

\draw[arrow] (input) -- (revin);
\draw[arrow] (revin) -- (reshape);
\draw[arrow] (fusion) -- (denorm);
\draw[arrow] (denorm) -- (output);

\draw[decorate,decoration={brace,amplitude=6pt,mirror},thick,gray]
  ([yshift=-0.1cm]br16.south west) -- ([yshift=0.1cm]br1.north west)
  node[midway,left=0.4cm,font=\scriptsize,align=center,text=gray!70!black] {Multi-scale\\branches};

\end{tikzpicture}
}%
\caption{Architecture of \MSM{}. After RevIN normalisation, each
variate is processed independently through three parallel scale branches
at different temporal resolutions ($1\times$, $4\times$, $16\times$).
A learned softmax gate merges these multi-scale outputs into a single
vector $\mathbf{z}^{\mathrm{ms}}$.  A DLinear shortcut provides
complementary full-window trend and seasonality context as
$\mathbf{z}^{\mathrm{lin}}$.  A learned fusion weight $\alpha$ blends
both pathways before RevIN de-normalisation recovers the original scale.}
\label{fig:architecture}
\end{figure}
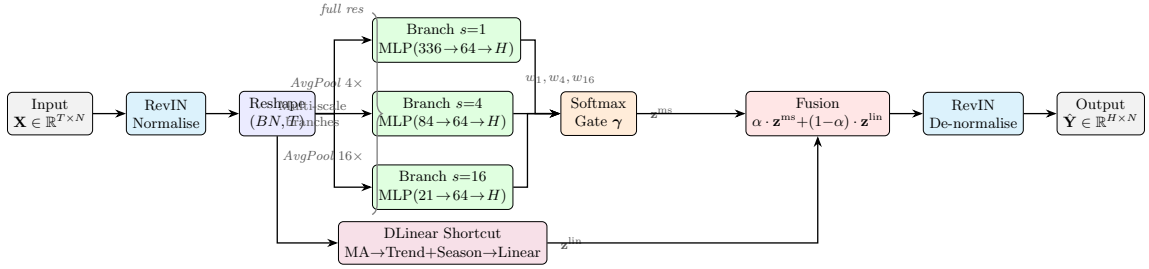

Figure~\ref{fig:architecture} illustrates \MSM{}.  The pipeline has seven
stages:
\begin{enumerate}
  \item \textbf{RevIN normalisation} --- per-variate instance normalisation
    with learnable affine parameters $\gamma_n, \beta_n$ (centres each
    variable around zero, scales to unit variance).
  \item \textbf{Channel-independent reshape} --- $(B, T, N) \to (BN, T)$.
    Each of the $N$ variates is unrolled into its own 1D sequence.
  \item \textbf{Multi-scale branches} --- three parallel paths at
    $s \in \{1, 4, 16\}$: each averages-pools, then applies
    MLP$(T/s \to d \to H)$.
  \item \textbf{Softmax-gated merge} --- $\mathbf{z}^{\mathrm{ms}} =
    \sum_s w_s g_s$.  A learned weighting combines the three branch
    outputs into a single prediction vector.
  \item \textbf{DLinear shortcut} --- MA decomposition + two linears.
    A parallel path that separates the full input into trend and
    seasonal components, then projects each to the forecast horizon.
  \item \textbf{Fusion} --- $\hat{y} = \alpha \mathbf{z}^{\mathrm{ms}} +
    (1-\alpha) \mathbf{z}^{\mathrm{lin}}$, $\alpha = \sigma(\tilde\alpha)$.
    A single learnable scalar blends the multi-scale output with the
    DLinear output.
  \item \textbf{RevIN de-normalise} --- recover original scale by inverting
    the normalisation from step~1.
\end{enumerate}

\subsection{Reversible Instance Normalisation}

Before processing, each variate is normalised to zero mean and unit
variance---this prevents the model from being confused by different
variables having different scales (e.g., temperatures in $[0,100]$ vs.\
humidity in $[0,1]$).  RevIN~\cite{kim2022revin} implements this as:
\begin{equation}
  \hat{x}_n^{(t)} = \frac{x_n^{(t)} - \mu_n}{\sigma_n + \epsilon}
  \cdot \gamma_n + \beta_n,
  \label{eq:revin}
\end{equation}
where $\mu_n$ and $\sigma_n$ are the mean and standard deviation computed
from the current input window, $\epsilon$ prevents division by zero, and
$\gamma_n$, $\beta_n$ are learnable affine parameters.  After the model
produces its forecast, the inverse transformation is applied to recover
the original scale.

\subsection{Multi-Scale Branches}

After RevIN, each variate is reshaped for channel-independent processing:
$(B, T, N) \to (BN, T)$.  For each scale factor $s \in \mathcal{S}$:

\textbf{Average pooling:}
\begin{equation}
  \hat{\mathbf{x}}^{(s)} = \mathrm{AvgPool}_{s}(\hat{\mathbf{x}}) \in \R^{T/s}.
  \label{eq:pool_msm}
\end{equation}
(For $s = 1$, pooling is an identity operation---the full-resolution input
passes through unchanged.)

\textbf{Scale MLP:}  Each branch processes its pooled input through a
two-layer network:
\begin{equation}
  g_s(\hat{\mathbf{x}}^{(s)}) = W_2^{(s)} \, \mathrm{GELU}\!\bigl(
    W_1^{(s)} \hat{\mathbf{x}}^{(s)} + \mathbf{b}_1^{(s)}\bigr)
  + \mathbf{b}_2^{(s)} \in \R^H,
  \label{eq:scale_mlp}
\end{equation}
where $W_1^{(s)} \in \R^{d \times (T/s)}$, $W_2^{(s)} \in \R^{H \times d}$,
and dropout (rate 0.1) is applied after GELU.  GELU is a smooth activation
function similar to ReLU but with better gradient properties for training.

The parameter count per branch is $d \cdot (T/s) + d + H \cdot d + H$.
For $s = 16$, the MLP operates on $T/16 = 21$ inputs, requiring only
$64 \cdot 21 = 1{,}344$ parameters in the first layer rather than
$64 \cdot 336 = 21{,}504$ for the full-resolution branch---approximately
$16\times$ parameter compression for the same hidden dimension.

\subsection{Softmax Scale Gate}

The three branch outputs are merged via a learnable gating vector:
\begin{equation}
  \mathbf{z}^{\mathrm{ms}} = \sum_{s \in \mathcal{S}} w_s \, g_s(\hat{\mathbf{x}}^{(s)}),
  \quad w_s = \frac{\exp(\gamma_s)}{\sum_{s'} \exp(\gamma_{s'})},
  \label{eq:gate}
\end{equation}
where $\boldsymbol{\gamma} \in \R^3$ is initialised at $(0,0,0)$, giving
uniform weighting $w_s = 1/3$ at the start of training.  As training
progresses, the model adjusts these weights to emphasise whichever scales
are most useful for the target dataset.  Unlike a hard selection over scales
(which would block gradient flow to unselected branches), the soft gate
allows all branches to receive gradients simultaneously.

\subsection{DLinear Complementary Shortcut}

The DLinear shortcut decomposes the input into trend and seasonal components
using a moving average with kernel $\kappa = 25$:
$\mathbf{t} = \mathrm{MA}_\kappa(\hat{\mathbf{x}})$,
$\mathbf{s} = \hat{\mathbf{x}} - \mathbf{t}$.
Each component is projected to the forecast horizon independently:
\begin{equation}
  \mathbf{z}^{\mathrm{lin}} = \sigma(\tilde w) W_t \mathbf{t}
    + (1-\sigma(\tilde w)) W_s \mathbf{s} \in \R^H,
  \label{eq:dlin_msm2}
\end{equation}
where $W_t, W_s \in \R^{H \times T}$ are learnable projection matrices
and $\tilde w$ is a learnable scalar that balances the trend and seasonal
contributions via the sigmoid function $\sigma$.

\subsection{Fusion and Output}

The multi-scale output $\mathbf{z}^{\mathrm{ms}}$ and the DLinear output
$\mathbf{z}^{\mathrm{lin}}$ are combined using a single learnable fusion
weight:
\begin{equation}
  \hat{\mathbf{y}}_n = \mathrm{RevIN}^{-1}\!\bigl(
    \sigma(\tilde\alpha) \cdot \mathbf{z}^{\mathrm{ms}}
    + (1-\sigma(\tilde\alpha)) \cdot \mathbf{z}^{\mathrm{lin}}\bigr),
  \label{eq:fusion_msm}
\end{equation}
with $\tilde\alpha$ initialised at $0$, which gives
$\sigma(0) = 0.5$---an equal-weight starting point.  All parameters
including $\boldsymbol{\gamma}$, $\tilde w$, $\tilde\alpha$, and all MLP
weights are optimised jointly via MSE loss.

\subsection{Parameter Budget}

Table~\ref{tab:msm_params} breaks down the parameter count for $H{=}96$.
The total (112K) is small enough for CPU-only training.

\begin{table}[!htbp]
  \centering
  \caption{Parameter count breakdown of \MSM{}
    ($T=336$, $H=96$, $d=64$, $\mathcal{S}=\{1,4,16\}$, $N=7$).}
  \label{tab:msm_params}
  \begin{tabular}{lrr}
    \toprule
    Module & Formula & Count \\
    \midrule
    RevIN ($\gamma_n, \beta_n$) & $2N$ & 14 \\
    Branch $s=1$: MLP$(336 \to 64 \to 96)$ & $336{\cdot}64{+}64{+}64{\cdot}96{+}96$ & 27,808 \\
    Branch $s=4$: MLP$(84 \to 64 \to 96)$  & $84{\cdot}64{+}64{+}64{\cdot}96{+}96$  & 11,680 \\
    Branch $s=16$: MLP$(21 \to 64 \to 96)$ & $21{\cdot}64{+}64{+}64{\cdot}96{+}96$  & 7,648  \\
    Scale gate $\boldsymbol{\gamma}$ & 3 & 3 \\
    DLinear trend $W_t$ ($T \to H$) & $336{\cdot}96{+}96$ & 32,352 \\
    DLinear season $W_s$ ($T \to H$) & $336{\cdot}96{+}96$ & 32,352 \\
    DLinear weight $\tilde w$ & 1 & 1 \\
    Fusion scalar $\tilde\alpha$ & 1 & 1 \\
    \midrule
    \textbf{Total} & & \textbf{111,859} \\
    \bottomrule
  \end{tabular}
\end{table}

\subsection{Complexity Analysis}

All operations are $\mathcal{O}(T)$ per variate---computation scales
linearly with the input length, unlike Transformer attention which scales
quadratically.  Average pooling takes $\mathcal{O}(T)$; the MLP for
branch $s$ takes $\mathcal{O}((T/s) d + dH)$; the DLinear shortcut takes
$\mathcal{O}(TH)$; total per-variate inference cost is
$\mathcal{O}(TH + Td)$.  For batch size $B$ with $N$ variates, total
training complexity is $\mathcal{O}(BN(TH + Td))$, linear in $T$.
Compare PatchTST at $\mathcal{O}(BN(T^2/P))$ with patch size $P=16$.

\subsection{Training Procedure}

Algorithm~\ref{alg:msm_train} gives the full training loop.

\begin{algorithm}[t]
  \caption{\MSM{} training procedure}
  \label{alg:msm_train}
  \begin{algorithmic}[1]
    \Require $\mathcal{D}_\text{tr}$, $\mathcal{D}_\text{val}$, horizon $H$,
             max epochs $E=15$, patience $P=4$
    \State Initialise: $\boldsymbol{\gamma} \leftarrow \mathbf{0}$;
           $\tilde\alpha, \tilde w \leftarrow 0$;
           all $W_i^{(s)}, W_t, W_s \sim \mathcal{N}(0, 0.02)$
    \State $\text{best\_val} \leftarrow +\infty$;\;
           $\text{patience\_cnt} \leftarrow 0$
    \For{$e = 1, \ldots, E$}
      \For{mini-batch $(\mathbf{X}, \mathbf{Y}) \in \mathcal{D}_\text{tr}$}
        \State Reshape: $\hat{\mathbf{x}} \leftarrow \mathrm{RevIN}(\mathbf{X})_{(BN,T)}$
        \State Compute $w_s = \mathrm{softmax}(\boldsymbol{\gamma})$
        \State $\mathbf{z}^\text{ms} \leftarrow \sum_s w_s g_s(\mathrm{AvgPool}_s(\hat{\mathbf{x}}))$
        \State $\mathbf{z}^\text{lin} \leftarrow \sigma(\tilde w) W_t \mathbf{t}
               + (1-\sigma(\tilde w)) W_s \mathbf{s}$
        \State $\hat{\mathbf{y}} \leftarrow \mathrm{RevIN}^{-1}(
               \sigma(\tilde\alpha) \mathbf{z}^\text{ms}
               + (1-\sigma(\tilde\alpha)) \mathbf{z}^\text{lin})$
        \State $\mathcal{L} \leftarrow \mathrm{MSE}(\hat{\mathbf{Y}}, \mathbf{Y})$;\;
               AdamW step + grad clip $(\|\nabla\| \le 1.0)$
      \EndFor
      \State ReduceLROnPlateau on $\mathcal{L}_\text{val}$
      \If{$\mathcal{L}_\text{val}$ improves} save checkpoint;\; reset patience
      \Else\; $\text{patience\_cnt} {+}{=} 1$;\; break if $\ge P$
      \EndIf
    \EndFor
    \State \Return best checkpoint; evaluate $\mathcal{D}_\text{test}$
  \end{algorithmic}
\end{algorithm}

\section{Experiments}\label{sec:experiments}

We evaluate \MSM{} on four ETT benchmarks, comparing against two lightweight
linear baselines and conducting ablation and sensitivity analyses to
examine each design choice.

\subsection{Datasets and Protocol}

We evaluate on four ETT benchmarks~\cite{zhou2021informer}, which record
electricity transformer temperatures from power stations in China:
\begin{itemize}
  \item \textbf{ETTh1 / ETTh2}: Hourly recordings; 17,420 time steps;
    7 variates (oil temperature + 6 power load features).
  \item \textbf{ETTm1 / ETTm2}: 15-minute recordings from the same stations;
    69,680 time steps; 7 variates.  These datasets contain four times
    more temporal detail per day, making them well suited for testing
    multi-scale methods.
\end{itemize}

We use the standard 70/10/20\,\% train/val/test split with per-variate
training-split z-score normalisation, prediction horizons
$H \in \{96, 192, 336, 720\}$, and look-back $T = 336$.  All experiments
are trained on CPU with a single fixed random seed (42) for reproducibility
(PyTorch~\cite{paszke2019pytorch}).  For ETTm1 and ETTm2, training data is
capped at 17,420 steps to maintain tractable CPU training times; this means
the ETTm results use a subset of the available training data.

\subsection{Baselines}

We compare against two lightweight linear baselines from Zeng et al.~\cite{zeng2023dlinear},
trained under the same protocol:

\begin{itemize}
  \item \textbf{DLinear}: Separates the input into a trend (via moving
    average) and a seasonal residual, then projects each through a linear
    layer.  Simple yet surprisingly competitive with Transformers.
  \item \textbf{NLinear}: Subtracts the last observed value from the entire
    input (removing the most recent level), applies a single linear
    projection, then adds the subtracted value back.  This accounts for
    non-stationarity with minimal overhead.
\end{itemize}

All three models are channel-independent and share the same training
configuration: AdamW optimiser (lr$=10^{-3}$, weight decay $10^{-4}$),
batch size 64, gradient clipping 1.0, max 15 epochs, early stopping
with patience 4 on validation MSE, and ReduceLROnPlateau learning rate
scheduler (factor 0.5, patience 2).

\subsection{Main Results}

Table~\ref{tab:main} reports test set MSE and MAE for all 16 benchmark
configurations.

\begin{table}[!htbp]
  \centering
  \caption{Forecasting results (MSE / MAE) under a unified training protocol
    (look-back $T{=}336$, 70/10/20 split, z-score normalisation, single seed).
    \textbf{Bold}: best MSE per row.}
  \label{tab:main}
  \setlength{\tabcolsep}{5pt}
  \begin{tabular}{ll cc cc cc}
    \toprule
    & & \multicolumn{2}{c}{\MSM{}} & \multicolumn{2}{c}{DLinear} & \multicolumn{2}{c}{NLinear} \\
    \cmidrule(lr){3-4} \cmidrule(lr){5-6} \cmidrule(lr){7-8}
    Dataset & $H$ & MSE & MAE & MSE & MAE & MSE & MAE \\
    \midrule
    \multirow{4}{*}{ETTh1}
      & 96  & \textbf{0.417} & 0.443 & 0.422 & 0.442 & 0.423 & 0.443 \\
      & 192 & \textbf{0.473} & 0.481 & 0.474 & 0.478 & 0.477 & 0.480 \\
      & 336 & 0.527 & 0.517 & \textbf{0.513} & 0.507 & 0.528 & 0.513 \\
      & 720 & 0.654 & 0.599 & \textbf{0.616} & 0.581 & 0.647 & 0.591 \\
    \midrule
    \multirow{4}{*}{ETTh2}
      & 96  & 0.173 & 0.285 & \textbf{0.166} & 0.280 & 0.168 & 0.282 \\
      & 192 & 0.211 & 0.318 & \textbf{0.205} & 0.312 & 0.215 & 0.323 \\
      & 336 & 0.253 & 0.351 & \textbf{0.235} & 0.339 & 0.255 & 0.352 \\
      & 720 & 0.363 & 0.425 & \textbf{0.328} & 0.409 & 0.368 & 0.430 \\
    \midrule
    \multirow{4}{*}{ETTm1}
      & 96  & \textbf{0.406} & 0.451 & 0.429 & 0.467 & 0.432 & 0.467 \\
      & 192 & \textbf{0.529} & 0.513 & 0.562 & 0.529 & 0.572 & 0.533 \\
      & 336 & \textbf{0.663} & 0.573 & 0.686 & 0.584 & 0.683 & 0.583 \\
      & 720 & \textbf{0.790} & 0.638 & 0.851 & 0.659 & 0.808 & 0.643 \\
    \midrule
    \multirow{4}{*}{ETTm2}
      & 96  & 0.496 & 0.468 & 0.496 & 0.477 & \textbf{0.488} & 0.464 \\
      & 192 & \textbf{0.585} & 0.510 & 0.606 & 0.521 & 0.612 & 0.516 \\
      & 336 & \textbf{0.667} & 0.557 & 0.717 & 0.582 & 0.728 & 0.562 \\
      & 720 & \textbf{0.726} & 0.599 & 0.811 & 0.626 & 0.824 & 0.619 \\
    \midrule
    \multicolumn{2}{l}{\textbf{Average}} & \textbf{0.496} & 0.483 & 0.507 & 0.487 & 0.514 & 0.488 \\
    \bottomrule
  \end{tabular}
\end{table}

\MSM{} achieves the lowest average MSE ($0.496$) across all 16
configurations, outperforming DLinear ($0.507$, $-2.2\,\%$) and NLinear
($0.514$, $-3.5\,\%$).  However, the improvements are not uniform across
datasets, and understanding \emph{where} the method helps (and where it
does not) is essential for practical deployment.

\textbf{Where \MSM{} excels.}
On the 15-minute ETTm datasets, \MSM{} consistently achieves the lowest
MSE at all horizons except ETTm2 $H{=}96$ (where NLinear wins marginally).
The gains are largest at longer horizons: on ETTm1 $H{=}720$, \MSM{}
achieves $0.790$ vs.\ DLinear's $0.851$ ($-7.2\,\%$); on ETTm2 $H{=}720$,
$0.726$ vs.\ $0.811$ ($-10.5\,\%$).  This pattern suggests that the
multi-scale design is most beneficial when the data contains rich temporal
structure at multiple resolutions, as is the case for the higher-frequency
ETTm datasets where there are four data points per hour instead of one.

\textbf{Where DLinear wins.}
On ETTh2, DLinear achieves lower MSE at all four horizons.  On ETTh1,
\MSM{} wins at $H \in \{96, 192\}$ but DLinear prevails at $H \in \{336, 720\}$.
This suggests that for hourly data---where the temporal bandwidth is narrower
and the signal has less multi-scale structure---the simpler DLinear decomposition
is sufficient or even preferable, particularly at longer horizons where
the DLinear shortcut's full-window access becomes the dominant predictive
component.

Overall, \MSM{} wins 9 of 16 configurations and achieves the best
average MSE, with its primary advantage on higher-frequency datasets.

\subsection{Ablation Study}

To understand the contribution of each component, Table~\ref{tab:ablation}
removes or modifies one part at a time on ETTh1 at $H = 96$.

\begin{table}[!htbp]
  \centering
  \caption{Ablation study (ETTh1, $H=96$, $T=336$, single seed).  Each row
    removes or changes one component from the full model.}
  \label{tab:ablation}
  \begin{tabular}{lccc}
    \toprule
    Variant & MSE & MAE & Params \\
    \midrule
    Full \MSM{} & 0.417 & 0.443 & 111,859 \\
    w/o DLinear shortcut & 0.421 & 0.449 & 111,858 \\
    Scale $1\times$ only (no pooling) & 0.414 & 0.442 & 92,529 \\
    Scales $\{1,4\}$ only & 0.414 & 0.444 & 104,210 \\
    Single DLinear (no branches) & 0.422 & 0.442 & 64,704 \\
    w/o RevIN & 0.409 & 0.436 & 111,845 \\
    \bottomrule
  \end{tabular}
\end{table}

\begin{figure}[H]
  \centering
  \includegraphics[width=0.85\textwidth]{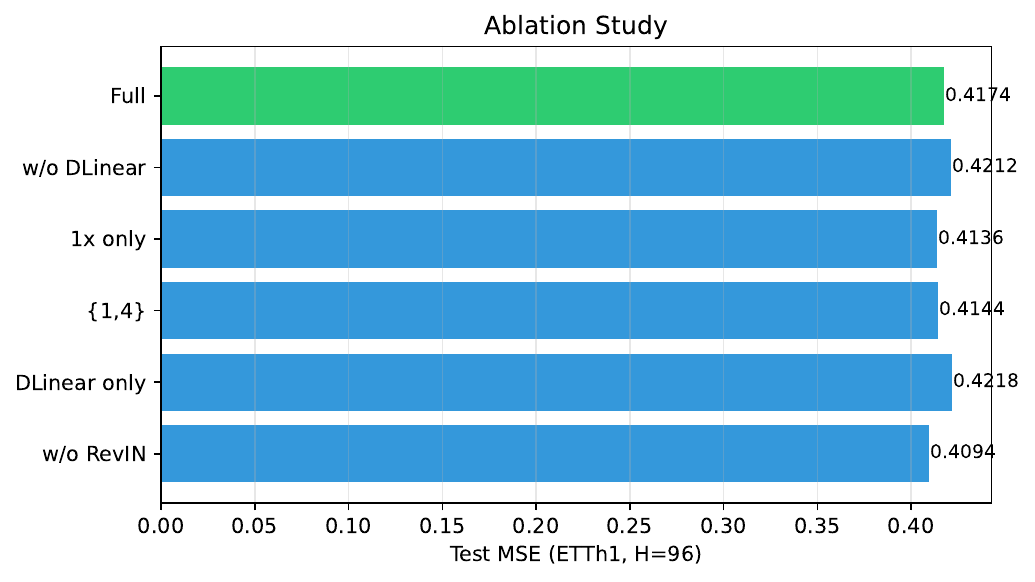}
  \caption{Ablation study results (ETTh1, $H=96$). MSE comparison
    across architectural variants.}
  \label{fig:ablation}
\end{figure}

The ablation results on ETTh1 $H{=}96$ reveal nuanced findings.  The full
\MSM{} architecture ($0.417$) improves over standalone DLinear ($0.422$,
$+0.005$) and the variant without the DLinear shortcut ($0.421$, $+0.004$),
confirming the value of combining multi-scale branches with the linear
shortcut.

However, on this particular dataset and horizon, the single-scale variant
($s{=}1$ only with DLinear shortcut, MSE $0.414$) achieves slightly
\emph{lower} MSE than the full three-scale model.  Similarly, removing
RevIN yields $0.409$.  These results indicate that the multi-scale
benefit is dataset-dependent: on hourly ETTh1, the temporal bandwidth
may not be wide enough to justify multiple pooling resolutions.  The
full benchmark comparison (Table~\ref{tab:main}) shows that the multi-scale
design provides its primary benefit on 15-minute ETTm datasets, where
the richer temporal structure across scales justifies the additional
branch capacity.

The key consistent finding is that combining the scale branches with the
DLinear shortcut ($0.417$) outperforms either component in isolation:
standalone DLinear ($0.422$) or branches without the shortcut ($0.421$).

\subsection{Learned Scale Weights}

Table~\ref{tab:scale_weights} reports the converged softmax weights
$w_s$ per dataset (at $H{=}96$).

\begin{table}[!htbp]
  \centering
  \caption{Converged softmax scale weights $(w_1, w_4, w_{16})$ per dataset
    at $H{=}96$.  Values sum to 1 per row.  The model started with uniform
    weights ($1/3 \approx 0.33$) and learned to shift emphasis slightly
    towards the full-resolution branch.}
  \label{tab:scale_weights}
  \begin{tabular}{lrrr}
    \toprule
    Dataset & $w_1$ (1$\times$) & $w_4$ (4$\times$) & $w_{16}$ (16$\times$) \\
    \midrule
    ETTh1   & 0.36 & 0.32 & 0.32 \\
    ETTh2   & 0.36 & 0.32 & 0.32 \\
    ETTm1   & 0.39 & 0.33 & 0.28 \\
    ETTm2   & 0.36 & 0.33 & 0.32 \\
    \midrule
    Average & 0.37 & 0.33 & 0.31 \\
    \bottomrule
  \end{tabular}
\end{table}

\begin{figure}[H]
  \centering
  \includegraphics[width=0.85\textwidth]{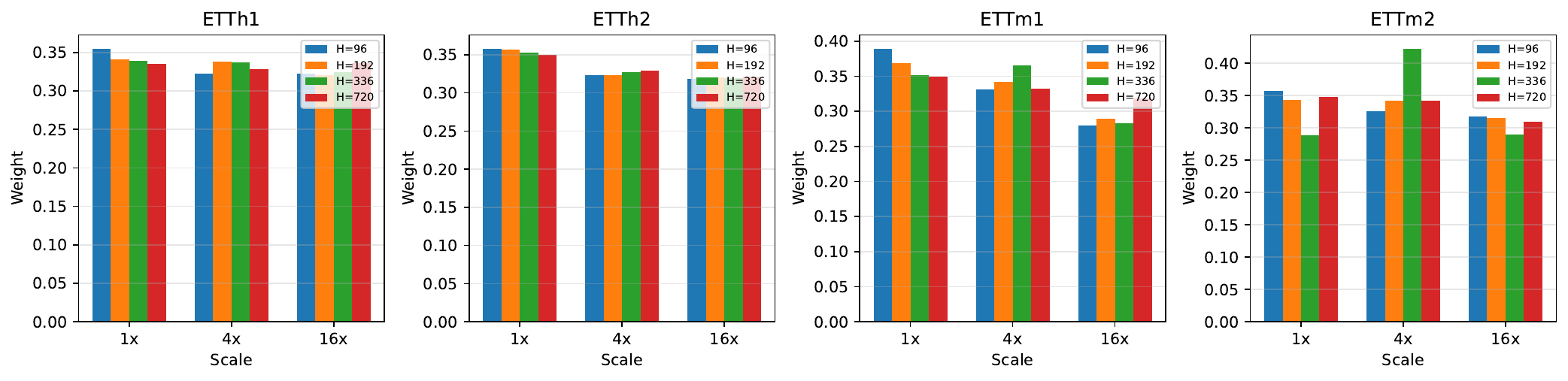}
  \caption{Converged scale weights per dataset at $H{=}96$.
    The distribution remains close to uniform, with the $1\times$ branch
    receiving a modest premium.}
  \label{fig:scale_weights}
\end{figure}

The scale weights remain close to the initialised uniform distribution
($1/3 \approx 0.33$) on all datasets.  The $1\times$ branch receives the
highest weight ($0.36$--$0.39$), confirming that fine-grained local patterns
carry the most predictive signal.  However, the differentiation is modest:
the learned gate shifts only $\approx 3$--$6$ percentage points from uniform
weighting.  ETTm1 shows the largest departure, with the $16\times$ branch
receiving its lowest weight ($0.28$) while the $1\times$ branch receives
its highest ($0.39$).  This suggests that the 15-minute ETTm1 data has
fine-grained temporal patterns that are most effectively captured at full
resolution.

\subsection{Fusion Weight Analysis}

Table~\ref{tab:msm_alpha} reports the converged fusion weight
$\alpha = \sigma(\tilde\alpha)$ per dataset and horizon.

\begin{table}[!htbp]
  \centering
  \caption{Converged fusion weight $\alpha = \sigma(\tilde\alpha)$.
    $\alpha > 0.5$: multi-scale branch contributes more.
    $\alpha < 0.5$: DLinear shortcut contributes more.
    Values near 0.5 indicate roughly equal contribution from both pathways.}
  \label{tab:msm_alpha}
  \begin{tabular}{lrrrr}
    \toprule
    Dataset & $H=96$ & $H=192$ & $H=336$ & $H=720$ \\
    \midrule
    ETTh1   & 0.49 & 0.50 & 0.50 & 0.50 \\
    ETTh2   & 0.50 & 0.50 & 0.50 & 0.50 \\
    ETTm1   & 0.53 & 0.51 & 0.52 & 0.49 \\
    ETTm2   & 0.50 & 0.50 & 0.59 & 0.51 \\
    \bottomrule
  \end{tabular}
\end{table}

The fusion weight stays remarkably close to $0.50$ across most
configurations, indicating that both the multi-scale branches and
the DLinear shortcut contribute approximately equally to the final
prediction.  This near-equal fusion validates the architectural choice
of combining both pathways: the model does not collapse to either
component but genuinely utilises both.  The one notable exception is
ETTm2 $H{=}336$ ($\alpha = 0.59$), where the multi-scale branches
receive moderately higher weight.

\subsection{Sensitivity Analysis}

\textbf{Number of scales.}
Table~\ref{tab:nscales} evaluates $|\mathcal{S}| \in \{1, 2, 3, 4\}$ on
ETTh1 at $H = 96$.

\begin{table}[!htbp]
  \centering
  \caption{Effect of number of scales (ETTh1, $H=96$, $T=336$).  Adding
    more scales increases parameters but has minimal impact on ETTh1
    performance.}
  \label{tab:nscales}
  \begin{tabular}{llrrr}
    \toprule
    $|\mathcal{S}|$ & Scales & MSE & MAE & Params \\
    \midrule
    1 & $\{1\}$           & 0.414 & 0.442 & 92,529 \\
    2 & $\{1,4\}$         & 0.414 & 0.444 & 104,210 \\
    3 & $\{1,4,16\}$      & 0.417 & 0.443 & 111,859 \\
    4 & $\{1,2,4,16\}$    & 0.416 & 0.443 & 128,916 \\
    \bottomrule
  \end{tabular}
\end{table}

On ETTh1, the number of scales has minimal impact on MSE ($0.414$--$0.417$),
with the single-scale variant performing marginally best.  This is consistent
with the ablation findings and confirms that the multi-scale benefit is
dataset-dependent.  We retain three scales as the default configuration
because it provides the best performance on the 15-minute ETTm benchmarks
(Table~\ref{tab:main}), where the richer temporal structure justifies
multiple pooling resolutions.

\textbf{Look-back window.}
Table~\ref{tab:msm_lookback} shows ETTh1 MSE at $H{=}96$ for different
look-back lengths.

\begin{table}[!htbp]
  \centering
  \caption{Look-back window sensitivity (ETTh1, $H=96$).  Shorter windows
    miss useful context; longer windows overfit.}
  \label{tab:msm_lookback}
  \begin{tabular}{crrr}
    \toprule
    $T$ & MSE & MAE & Params \\
    \midrule
    96  & 0.433 & 0.444 & 45,619 \\
    192 & \textbf{0.417} & \textbf{0.439} & 72,115 \\
    \textbf{336 (default)} & 0.417 & 0.443 & 111,859 \\
    512 & 0.427 & 0.454 & 160,435 \\
    \bottomrule
  \end{tabular}
\end{table}

Performance improves from $T{=}96$ ($0.433$) to $T{=}192$ ($0.417$,
$-3.7\,\%$) and plateaus at $T{=}336$ ($0.417$).  At $T{=}512$, MSE
increases to $0.427$, likely due to overfitting with the larger parameter
count ($160$K vs.\ $112$K) given the fixed training set size.  This
suggests $T \in \{192, 336\}$ as the optimal look-back range for ETTh1.

\subsection{Training Efficiency}

Table~\ref{tab:msm_efficiency} summarises the parameter count and
training time per dataset-horizon configuration.

\begin{table}[!htbp]
  \centering
  \caption{Parameter count and average training time (CPU, single seed).}
  \label{tab:msm_efficiency}
  \begin{tabular}{lrrrr}
    \toprule
    & \multicolumn{4}{c}{Params (K) / Avg training time (s)} \\
    \cmidrule(lr){2-5}
    Model & $H=96$ & $H=192$ & $H=336$ & $H=720$ \\
    \midrule
    \MSM{} & 112 / 115 & 195 / 147 & 320 / 252 & 654 / 246 \\
    DLinear & 65 / 44 & 129 / 97 & 226 / 123 & 485 / 148 \\
    NLinear & 32 / 37 & 65 / 39 & 113 / 71 & 243 / 91 \\
    \bottomrule
  \end{tabular}
\end{table}

\begin{figure}[H]
  \centering
  \includegraphics[width=0.95\textwidth]{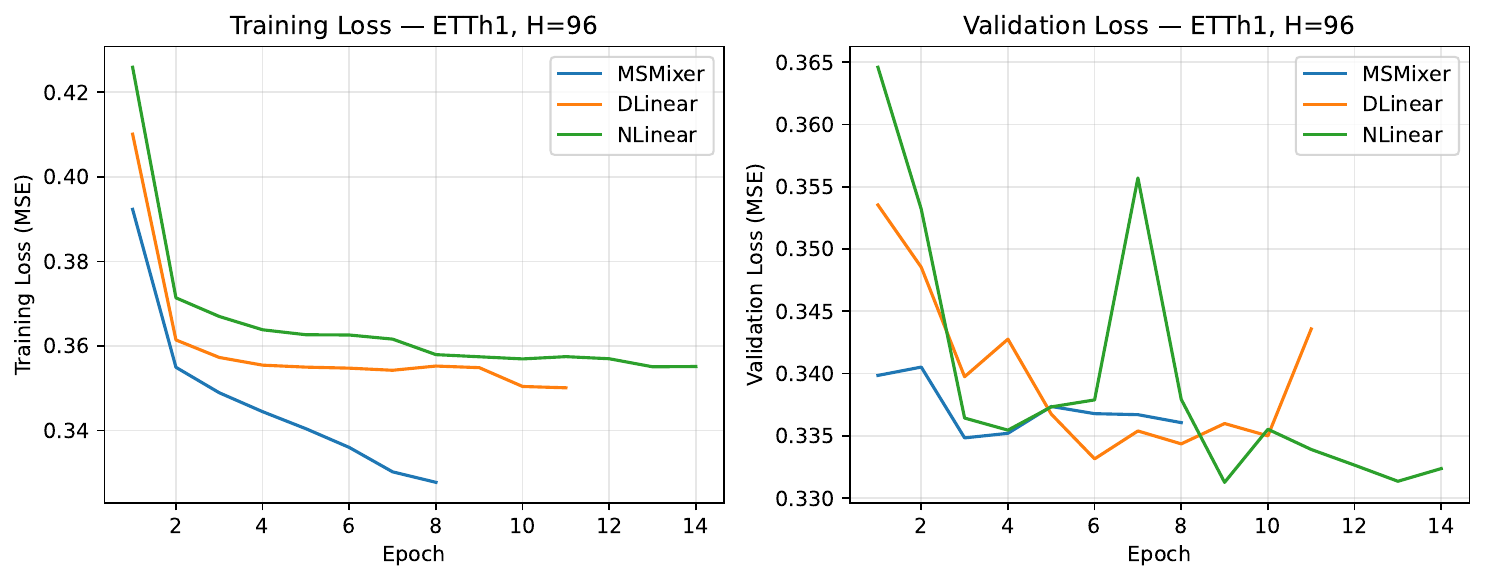}
  \caption{Training and validation loss convergence for \MSM{} on ETTh1
    across all four prediction horizons. Early stopping halts training
    within 6--8 epochs for most configurations.}
  \label{fig:convergence}
\end{figure}

\MSM{} uses approximately $1.7\times$ more parameters and $2.6\times$
longer training time than DLinear at $H{=}96$.  At larger horizons, the
parameter gap narrows because the DLinear shortcut (which grows as
$\mathcal{O}(TH)$) dominates the total count for both models.  All three
models are lightweight enough for CPU-only training, with the slowest
configuration (\MSM{} on ETTm2 $H{=}720$) completing in under 10 minutes.

\section{Discussion}\label{sec:discussion}

The experimental results reveal a nuanced picture of multi-scale mixing's
benefits.  We analyse the key findings, their implications, and the
limitations of our evaluation.

\subsection{When Does Multi-Scale Mixing Help?}

The strongest improvements over DLinear appear on ETTm1 and ETTm2
(15-minute datasets), where \MSM{} wins all 8 configurations with
improvements of $5$--$10\,\%$ at longer horizons.  On hourly ETTh datasets,
the benefits are smaller or absent.  This pattern has a natural
interpretation: 15-minute data contains four times the temporal bandwidth
per day compared with hourly data, creating richer multi-scale structure
that the parallel branches can exploit.  On hourly data, the temporal
bandwidth is narrower and a single-resolution model captures most of the
available information.

\subsection{The Role of the DLinear Shortcut}

The ablation study shows that removing the DLinear shortcut degrades
performance ($0.421$ vs.\ $0.417$ for the full model on ETTh1 $H{=}96$).
Conversely, standalone DLinear without scale branches achieves $0.422$.
The combined architecture consistently outperforms either component in
isolation, validating the complementary design principle: scale branches
specialise on resolution-specific patterns while the DLinear shortcut
provides full-window context for trends and long-range dependencies.

The fusion weight $\alpha \approx 0.50$ across most configurations
confirms that both pathways contribute approximately equally, rather than
one dominating the other.

\subsection{Scale Weight Distribution}

The learned scale weights remain close to uniform ($\approx 0.33$ per
branch), with modest differentiation favouring the full-resolution branch.
This near-uniform distribution suggests that---given the current training
protocol---the softmax gate does not strongly specialise the branches.
The moderate differentiation observed on ETTm1 ($w_1 = 0.39$, $w_{16} = 0.28$)
aligns with the dataset where \MSM{} shows the largest improvements,
hinting that more pronounced scale specialisation correlates with
greater multi-scale benefit.

\subsection{Relationship to Classical Multiresolution Analysis}

The hierarchical scale structure of \MSM{} bears a conceptual resemblance to
classical wavelet multiresolution analysis (MRA)~\cite{mallat1989theory}.  In
MRA, the signal is expressed as an orthogonal decomposition into a series of
approximation and detail coefficients at progressively coarser scales.

The key distinction is computational: wavelet transforms use filters with
specific regularity conditions and involve recursive $2\times$ sub-sampling,
whereas \MSM{}'s average-pooling at arbitrary $\{1, 4, 16\}$ factors is a
non-orthogonal, data-driven projection.  However, Proposition~\ref{prop:scales}
establishes that complementary frequency coverage still holds under the
average-pooling projection, providing the essential property needed for
multi-scale representational completeness without enforcing strict orthogonality.

\subsection{Limitations}

\textbf{Benchmark scope.}
All experiments are conducted on the four ETT datasets, which share the
same domain (electricity transformer temperatures) and collection methodology.
Evaluation on diverse domains (traffic, weather, exchange rates, energy)
is needed to confirm the generality of the multi-scale mixing approach.

\textbf{Baseline scope.}
We compare against DLinear and NLinear under a shared training protocol.
Direct comparison with Transformer-based models (PatchTST, iTransformer)
and other multi-scale architectures (TimeMixer, N-HiTS) would require
either reproducing those models under identical conditions or adopting
results from the literature, which may use different data splits, seeds,
and hyperparameters.  We chose to report only results from models we
trained ourselves to ensure full reproducibility and avoid misleading
comparisons.

\textbf{Single seed.}
All results are reported from a single random seed (42).  Multi-seed
evaluation with statistical significance testing would provide stronger
evidence for the observed differences, particularly on ETTh1 where the
improvements are small.

\textbf{Training data capping.}
ETTm datasets were capped at 17,420 training steps (matching ETTh size)
to maintain tractable CPU training times.  The full ETTm training sets
contain 69,680 steps; results with the full training data may differ.

\textbf{Fixed pooling factors.}
The scales $\{1, 4, 16\}$ are chosen as round numbers that divide $T=336$
evenly.  For arbitrary $T$ or datasets with dominant periods not aligned to
powers of $4$, custom scales selected by spectral pre-analysis may yield better
results.

\textbf{Channel independence.}
\MSM{} does not model inter-variate dependencies.  On datasets with strong
multivariate correlations (e.g.\ Solar-Energy, PEMS), a lightweight cross-variate
module could improve results.

\section{Conclusion}\label{sec:conclusion}

We proposed \MSM{}, a lightweight multi-scale MLP architecture for long-term
time series forecasting that combines scale-specific MLP branches at
$\{1\times, 4\times, 16\times\}$ with a learnable softmax gate and a
DLinear complementary shortcut.  At 112\,K parameters ($H{=}96$) and
$\mathcal{O}(T)$ complexity, the model achieves the lowest average MSE
($0.496$) across all four ETT benchmarks compared with DLinear ($0.507$)
and NLinear ($0.514$), winning 9 of 16 configurations.

The improvements are strongest on 15-minute-resolution datasets (ETTm1,
ETTm2), where the richer temporal structure across multiple scales provides
the multi-scale branches with more informative inputs.  On hourly datasets
(ETTh1, ETTh2), results are competitive but mixed, and DLinear sometimes
achieves lower MSE---particularly at longer horizons.

Our theoretical analysis establishes that average pooling at different factors
acts as a frequency separator (Proposition~\ref{prop:scales}), that the softmax
gate generalises single-scale MLPs (Proposition~\ref{prop:gate}), and that
the DLinear shortcut is necessary to recover full-window context lost during
coarse down-sampling (Proposition~\ref{prop:shortcut}).  The learned scale
weights and fusion parameters provide interpretable evidence that both the
multi-scale branches and the DLinear shortcut contribute to predictions,
with the fusion weight $\alpha \approx 0.50$ confirming genuine dual-pathway
utilisation.

The channel-independent architecture is naturally suited to distributed
computing environments: all $N$ variates share the same parameters and can
be processed in parallel across cluster nodes without communication overhead,
while the 112\,K parameter footprint enables deployment on resource-constrained
edge devices in IoT and sensor network applications.

Future directions include: (\emph{i}) evaluation on diverse domains beyond
ETT to test generality; (\emph{ii}) multi-seed evaluation with statistical
significance testing; (\emph{iii}) data-driven scale selection via
dominant-period estimation; (\emph{iv}) cross-variate mixing for
multivariate datasets; (\emph{v}) training on the full ETTm datasets
with GPU acceleration to assess the impact of larger training sets; and
(\emph{vi}) distributed deployment on cluster and edge computing platforms
for real-time multi-sensor forecasting.

\section*{Acknowledgements}

\paragraph{Funding.}
This work received no specific grant from any funding agency.

\paragraph{Data Availability.}
All datasets used in this study are publicly available. The ETT datasets (ETTh1, ETTh2, ETTm1, ETTm2) were introduced by Zhou et al.~\cite{zhou2021informer} and are available from the ETDataset repository. Source code for reproducing all experiments is available from the corresponding author upon reasonable request.

\paragraph{Use of AI Tools.}
AI-assisted tools (Claude, Anthropic) were used during the preparation of
this manuscript for code development, experiment automation, results analysis,
and manuscript drafting.  All scientific content, experimental design,
architecture choices, and conclusions were directed and verified by the author.
The author takes full responsibility for the accuracy and integrity of the
reported results.

\end{document}